\pdfoutput=1
\documentclass{cambrian}

\usepackage{booktabs}       
\usepackage{amsfonts}       
\usepackage{nicefrac}       
\usepackage{microtype}      
\usepackage{multirow}       
\definecolor{bluelink}{RGB}{0,113,188}
\definecolor{greenlink}{RGB}{0,188,113}
\definecolor{redlink}{RGB}{188,37,37}
\hypersetup{
    colorlinks=true,%
    citecolor=green!90!black,%
    filecolor=redlink,%
    linkcolor=red!90!black,%
    urlcolor=bluelink
}

\usepackage[ruled,vlined,noend]{algorithm2e}
\usepackage{subcaption}
\usepackage{xspace}
\usepackage{wrapfig}

\usepackage{tikz}

\usepackage{titlesec}
\titlespacing*{\paragraph}
    {0pt}     
    {0.25em}  
    {0.25em}  

\newif\ifshowtodos
\showtodosfalse 
\ifshowtodos
    \newcommand{\todotxt}[1]{#1}
\else
    \newcommand{\todotxt}[1]{}
\fi

\definecolor{navyblue}{HTML}{0071BC}

\newcommand{\methodDiagnose}{TsT\xspace}
\newcommand{\methodDiagnoseLong}{{Test-set Stress-Test}\xspace}
\newcommand{\methodDebias}{IBP\xspace}
\newcommand{\methodDebiasLong}{{Iterative Bias Pruning}\xspace}

\definecolor{rowheader}{RGB}{240,240,240}
\definecolor{oursrowheader}{RGB}{200,220,255}
\definecolor{ours}{RGB}{220,240,255}
\definecolor{inccol}{RGB}{245,245,245}
\definecolor{inctext}{RGB}{100,100,100}
\definecolor{pos}{RGB}{0,128,0}
\definecolor{neg}{RGB}{200,0,0}
\definecolor{navyblue}{HTML}{0071BC}

\definecolor{ditblue}{HTML}{5D8DFD}
\definecolor{ditred}{HTML}{F5433D}

\usepackage[numbers,sort&compress]{natbib}
\usepackage{placeins}

\hypersetup{pdftitle={Benchmark Designers Should "Train on the Test Set" to Expose Exploitable Non-Visual Shortcuts},
    pdfauthor={Ellis Brown, Jihan Yang, Shusheng Yang, Rob Fergus, Saining Xie}}

\title{
    \centering
    \LARGE
    Benchmark Designers Should ``Train on the Test Set''\\
    to Expose Exploitable Non-Visual Shortcuts
    \vspace{-0.5em}
}

\author{
    Ellis Brown \quad
    Jihan Yang \quad
    Shusheng Yang \quad
    Rob Fergus \quad
    Saining Xie \\
    New York University
}

\begin{abstract}
    Multimodal LLMs can answer many questions in vision-centric benchmarks without looking at the image, using linguistic priors and skewed answer distributions.
    Blind evaluation shows what a model already answers this way, but not what it could learn from the test set's own regularities.
    Extending partial-input auditing (e.g., hypothesis-only baselines in natural language inference), we argue that benchmark designers should ``train on the test set'': probe the artifact they release for exploitable patterns.
    Our \methodDiagnoseLong{} (\methodDiagnose{}) cross-validates a text-only Qwen2-7B on the test set's questions and answer options, yielding a benchmark-level score and a per-question bias score $s(x)$; a random forest on hand-crafted features adds a fast, interpretable audit.
    On the template-based VSI-Bench and CV-Bench, the held-out score is 17.9 and 13.1 points above the model's own zero-shot score.
    On MMMU it learns little, even though GPT-4o correctly answers 52.3\% of its multiple-choice questions without the image, so blind success there reflects pretrained knowledge rather than learnable test-set patterns.
    \methodDebiasLong{} (\methodDebias{}) removes the questions with the highest $s(x)$ and re-diagnoses; on VSI-Bench it widens a fine-tuned model's vision-blind gap more than random removal at the same rate.
    We also release VSI-Bench-Debiased, which lowers a fine-tuned model's blind score from 44.7 to 32.0 while its vision score falls only from 57.1 to 48.7.
\end{abstract}

\renewcommand{\thefootnote}{\arabic{footnote}}  


\begin{document}
\maketitle
\par\vspace*{1em}\noindent\textbf{Project Page:}\quad \url{https://vision-x-nyu.github.io/test-set-training/}

{\let\thefootnote\relax\footnotetext{Published as a conference paper at COLM 2026.}}
\newpage

\section{Introduction}\label{sec:introduction}

\begin{figure}[t]
    \centering
    \includegraphics[width=\linewidth]{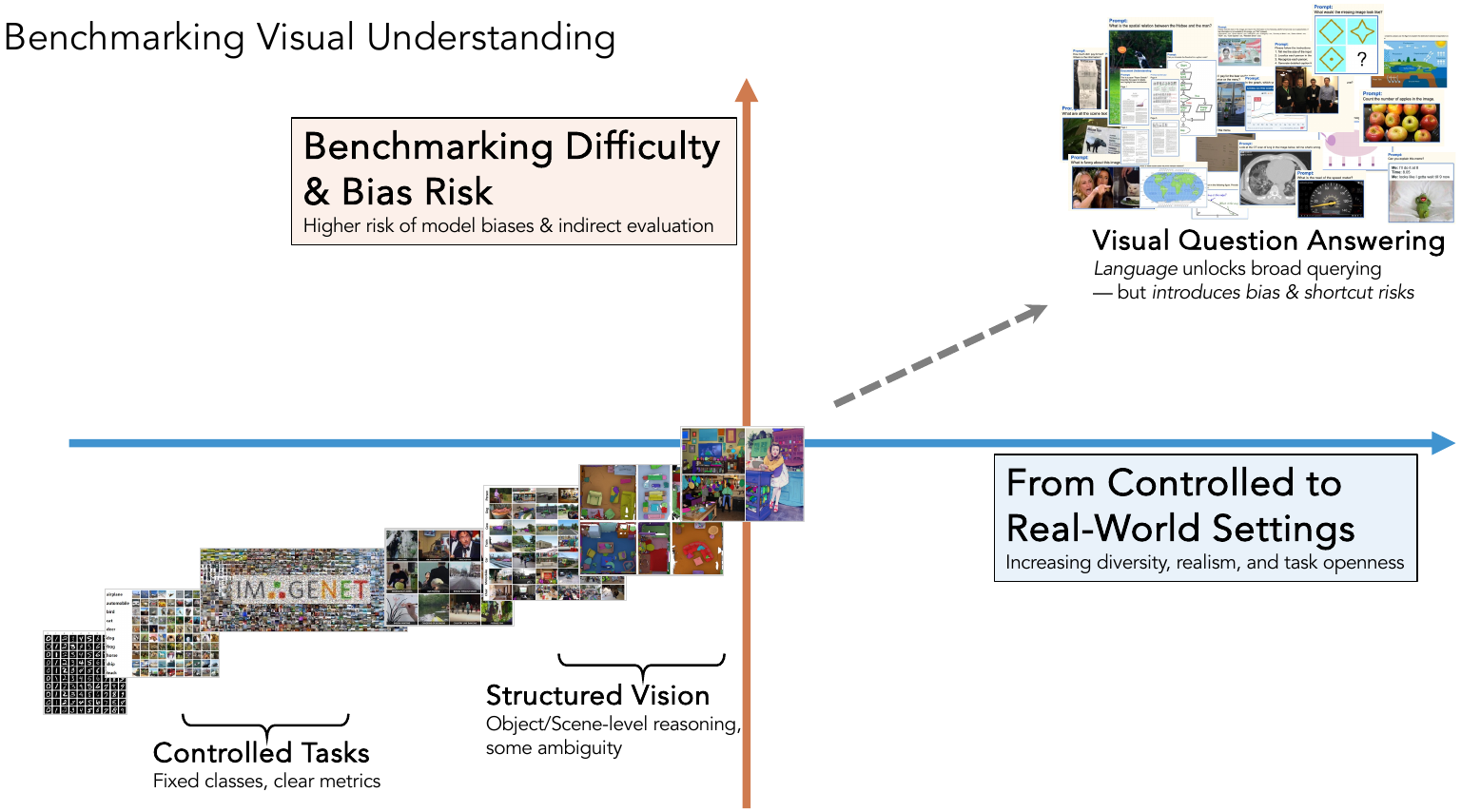}
    \caption{\textbf{Visual benchmarks have moved from controlled tasks to open-ended VQA.}
    Asking questions in language makes benchmarks more expressive, but it also exposes them to non-visual shortcuts: models can score well by exploiting linguistic patterns without looking at the image.}
    \label{fig:teaser}
\end{figure}

Benchmarks have driven progress in computer vision, from MNIST~\citep{lecun1998gradient} and ImageNet~\citep{deng2009imagenet} to COCO~\citep{lin2014microsoft} and VQA~\citep{antol2015vqa}. As models grew more capable, benchmarks moved from narrow, controlled tasks to open-ended, real-world evaluation (\cref{fig:teaser}). Multimodal LLMs (MLLMs) now let us ask anything about an image, but this flexibility has a cost: we have lost control over what is being measured.

A model can answer a visual question without looking at the image, using linguistic priors and skewed answer distributions. Language bias was already a concern in the VQA era (\cref{sec:related_work}); LLMs, as both knowledge engines and sources of shortcuts, make it pressing again~\citep{tong2024cambrian}.
The usual safeguard is the ``blind'' test, which evaluates a multimodal benchmark with vision disabled~\citep{antol2015vqa,goyal2017making,chen2024mmstar,tong2024cambrian}. It shows what a model can already answer without looking. It does not show which questions become answerable once a model learns the test set's own regularities, or how to fix them.

If a benchmark \emph{can} be gamed, it \emph{will} be. We argue that \textbf{benchmark designers should ``train on the test set''}\footnote{The phrasing is deliberately provocative and addressed to benchmark designers. We propose a held-out, pre-release audit of the benchmark itself, not a recipe for training deployed models on test answers.} to expose exploitable non-visual shortcuts: fit a probe to the test set's non-visual input and measure what it learns. As in adversarial filtering~\citep{zellers2018swag,lebras2020adversarial}, benchmark design becomes iterative and adversarial.

The probe audits the data, not a model. Trained on a separate in-distribution split, where one exists, it finds biases of the process that generated the data. Cross-validated within the released test set, it can also catch \emph{idiosyncratic vulnerabilities} of that specific artifact, such as those introduced by finite-sample selection, question templates, or human filtering.

Our \methodDiagnoseLong{} (TsT) applies $k$-fold cross-validation to the test set's non-visual input and yields a benchmark-level TsT score and a per-question bias score $s(x)$. TsT-LLM LoRA-tunes a language model on question and answer-option text; TsT-RF, a random forest on hand-crafted features, adds a fast, interpretable audit. \methodDebiasLong{} (IBP) then adapts AFLite's iterative filtering~\citep{lebras2020adversarial}: it repeatedly removes the questions with the highest $s(x)$ and re-diagnoses the rest.

Learnable shortcuts vary widely across benchmarks (\cref{sec:diagnosing_shortcuts}). On the template-based VSI-Bench~\citep{yang2024think} and CV-Bench~\citep{tong2024cambrian}, a blind Qwen2-7B cross-validated on the test set's text scores 17.9 and 13.1 points above its own zero-shot score. On knowledge-intensive MMMU~\citep{yue2023mmmu} it learns little, even though GPT-4o~\citep{hurst2024gpto} correctly answers 52.3\% of MMMU's multiple-choice questions without the image; blind success there points mostly to pretrained knowledge.

We compare every pruning method with random removal at the same rate, on VSI-Bench and on MMMU, a non-templated negative control (\cref{sec:mitigating_shortcuts}). On VSI-Bench, a designer-in-the-loop pilot, built with hand-written per-type filters rather than by IBP, sharply lowers the blind score of a model fine-tuned on in-distribution data. It also widens the model's vision-blind gap more than random removal does. Automated per-type IBP, run within each question type at the same removal rate, widens it less than the pilot but more than random removal. On MMMU, pruning behaves like random removal.

Our contributions are:
(1) a definition of an ``exploitable'' non-visual shortcut, with the designer deciding how to respond to flagged questions (\cref{sec:problem_statement});
(2) the TsT diagnostic, with its LLM and random forest variants (\cref{sec:diagnosing_shortcuts});
(3) diagnoses of VSI-Bench, CV-Bench, MMMU, and VideoMME, with a GPT-4o blind baseline (\cref{subsec:diagnostics_validation});
(4) IBP, which adapts AFLite's iterative filtering~\citep{lebras2020adversarial} to TsT scores, evaluated against random removal (\cref{sec:mitigating_shortcuts}); and
(5) the VSI-Bench pilot, released as VSI-Bench-Debiased (\cref{subsec:mitigation_vsi_robust}).

\section{The Challenge: Non-Visual Shortcuts}\label{sec:problem_statement}

We define an \textbf{exploitable non-visual shortcut} as any pattern, regardless of origin, that renders visual input redundant for a task designed to measure visual understanding.
Such shortcuts take two forms.

\begin{figure}[t]
    \centering
    \includegraphics[width=1\linewidth]{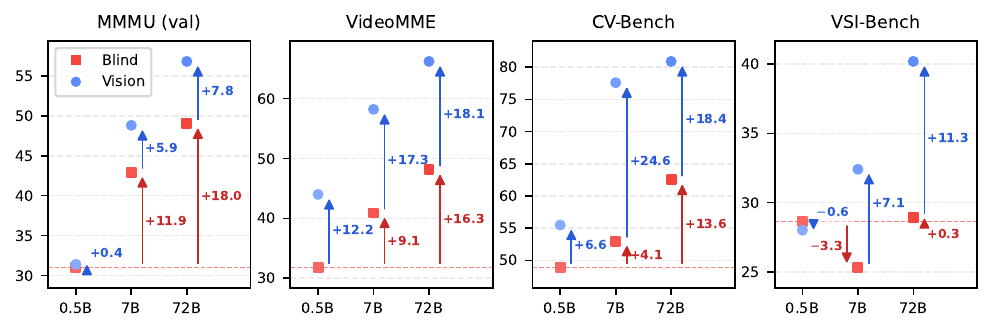}
    \vspace{-1.75em}
    \caption{
        \textbf{Blind scores rise with LLM scale on MMMU but not on VSI-Bench.}
        Blind
            \protect\tikz[baseline=0.05ex]{\protect\fill[fill={ditred}] (0,0) rectangle (1.75ex,1.75ex);}
        and vision-enabled
            \protect\tikz[baseline=-0.8ex]{\protect\fill[fill={ditblue}] (0,0) circle (0.9ex);}
        scores of LLaVA-OneVision~\citep{li2024llavaov} at three LLM sizes.
        Red arrows give the blind gain over the smallest model's blind score (dashed line); blue arrows, the gain from enabling vision.
        On MMMU, scaling the LLM adds more than enabling vision does; on VideoMME and CV-Bench, both help; on VSI-Bench, blind scores stay flat and the gains come from vision.
        VSI-Bench values are per-task averages from \citet{yang2024think}.
    }\label{fig:benchmark_lov}
\end{figure}

\emph{Knowledge-based} shortcuts come from what an LLM learned in pretraining~\citep{tong2024cambrian}: ``How tall is the refrigerator?'' can be answered from memorized appliance dimensions.
On MMMU, blind scores rise with the size of the LLM backbone, consistent with such shortcuts; on VSI-Bench, they do not (\cref{fig:benchmark_lov}).

\begin{figure}[t]
    \centering
    \begin{subfigure}[t]{0.49\linewidth}
        \centering
        \includegraphics[width=\linewidth]{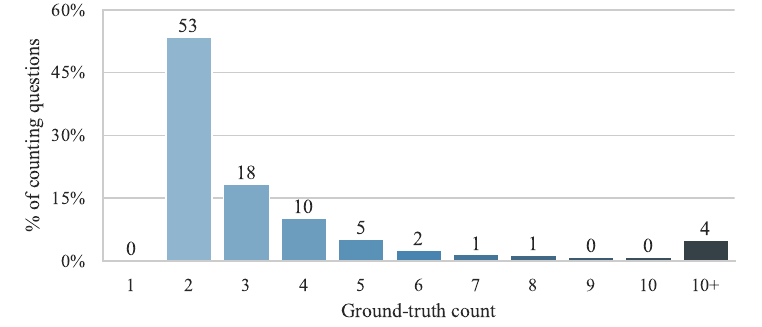}
        \caption{\small\centering Counting (VSI-Bench)}
    \end{subfigure}
    \hfill
    \begin{subfigure}[t]{0.49\linewidth}
        \centering
        \includegraphics[width=\linewidth]{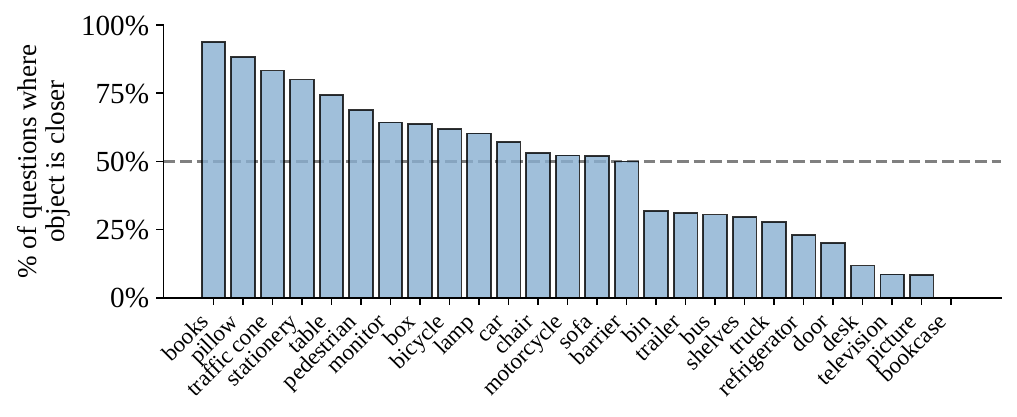}
        \caption{\small\centering Spatial relation (CV-Bench depth)}
    \end{subfigure}
    \vspace{0.5em}
    \begin{subfigure}[t]{0.49\linewidth}
        \centering
        \includegraphics[width=\linewidth]{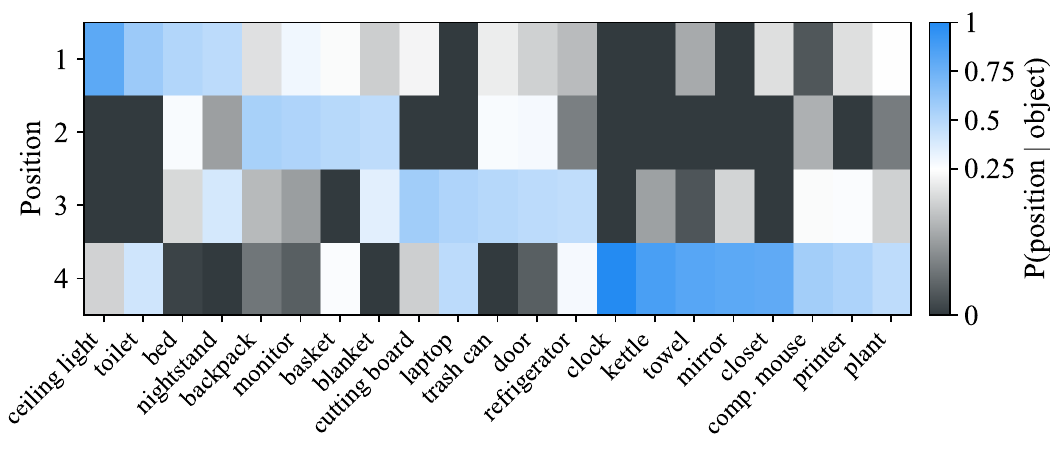}
        \caption{\small\centering Appearance order (VSI-Bench)}
    \end{subfigure}
    \hfill
    \begin{subfigure}[t]{0.49\linewidth}
        \centering
        \includegraphics[width=\linewidth]{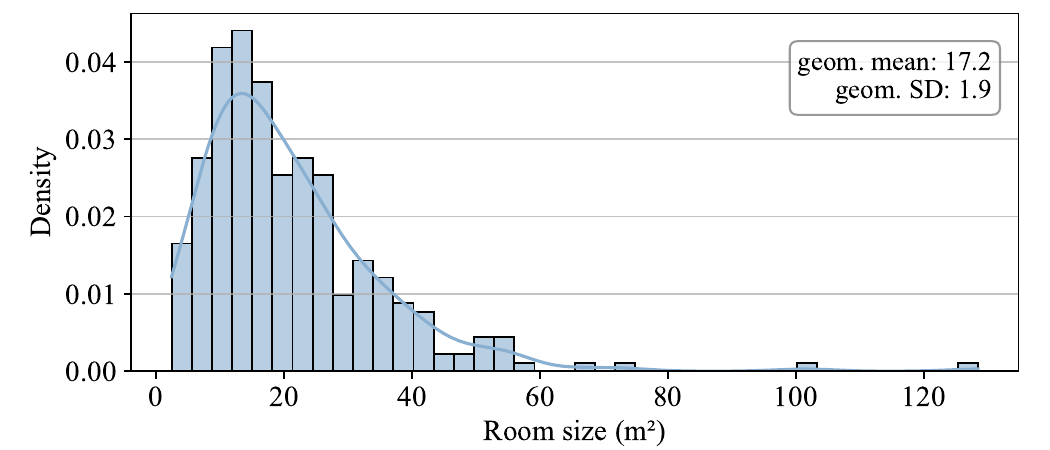}
        \caption{\small\centering Room size (VSI-Bench)}
    \end{subfigure}
    \caption{
        \textbf{Statistical biases create non-visual shortcuts.}
        (a) Share of counting questions with each gold count.
        (b) How often each object is the closer one, for objects in at least 10 questions (dashed line: chance).
        (c) How often each object occupies each position of the gold appearance order, partly reflecting questions that share a video.
        (d) Room-size distribution, with geometric mean and geometric standard deviation.
    }\label{fig:vsi_bias_examples}
    \vspace{-0.75em}
\end{figure}

\emph{Statistical} shortcuts come from regularities in a benchmark's questions and answers that a model can learn instead of visual reasoning (\cref{fig:vsi_bias_examples}); they can be strong even where knowledge-based shortcuts are weak, as on VSI-Bench.
Over half of VSI-Bench counting answers are ``2''.
In CV-Bench depth questions, books are the closer object in 60 of 64 questions, and a bookcase in none of 42.
In VSI-Bench, objects favor particular positions in appearance-order questions, and room sizes are roughly log-normal.

Training data produced by the same generator carries the same regularities, so models can pick up these shortcuts without access to the test set.
Fine-tuning LLaVA-Video-7B~\citep{zhang2024video} on VSI-Train-10k, 10K in-distribution questions generated with VSI-Bench's templates from training scenes (\cref{app:vsi-train-details}), raises its \emph{blind} VSI-Bench score by 18.8 points, nearly matching its 20.4-point gain with vision (\cref{tab:vsi_robust_results}, original columns).
A benchmark's exposure to such shortcuts is therefore worth measuring, for the benchmark as a whole and for each question.

TsT is meant for benchmark designers, before release.
A high bias score $s(x)$ (\cref{subsec:tst_framework}) flags a question as exploitable; the designer decides whether to diversify the answer distribution, rewrite the question, or accept the bias.

\section{Diagnosing Non-visual Shortcuts via Test-Set Stress-Testing}\label{sec:diagnosing_shortcuts}

Benchmarks can fail in two ways: \emph{training failures}, where biased training data prevents robust generalization (e.g., VQA-CP;~\citealp{agrawal2018don}), and \emph{evaluation failures}, where the test set itself contains exploitable artifacts.
TsT targets the latter.
A diagnostic trained on separate in-distribution data finds biases of the data-generating process but can miss idiosyncratic vulnerabilities of a particular test set (\cref{fig:tst_overview:bias_space}).
We therefore probe the test set itself, which is also the practical choice: most MLLM benchmarks are released with only an evaluation split.

\subsection{The TsT Framework}\label{subsec:tst_framework}

\begin{figure}[t]
    \centering
    \begin{subfigure}[b]{0.3925\columnwidth}
        \centering
        \includegraphics[width=0.95\linewidth]{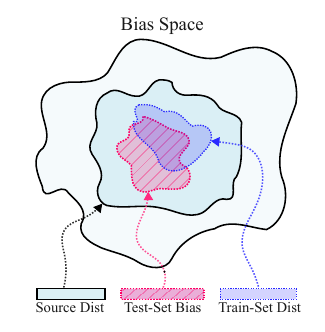}
        \caption{
            \small
            \centering Bias space
        }\label{fig:tst_overview:bias_space}
    \end{subfigure}
    \hfill
    \begin{subfigure}[b]{0.5975\columnwidth}
        \centering
        \includegraphics[width=0.95\linewidth]{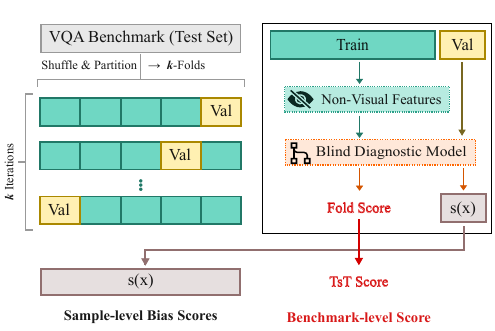}
        \caption{
            \small
            \centering Diagnostic pipeline
        }\label{fig:tst_overview:diagnostic_pipeline}
    \end{subfigure}
    \vspace{-0.25em}
    \caption{
        \textbf{TsT probes the test set directly.}
        (a) A diagnostic trained on separate data from the source distribution (dotted blue) finds only the test-set biases (pink) that its training data shares and can miss those specific to the test set.
        (b) The test set is split into $k$ folds; a blind diagnostic trains on $k{-}1$ folds and scores the held-out fold, $k$ times over, giving a benchmark-level TsT score and a per-question bias score $s(x)$.
    }\label{fig:tst_overview}
    \vspace{-0.75em}
\end{figure}

TsT applies $k$-fold cross-validation~\citep{stone1974cross,hastie2009elements} to the test set: split it into $k$ disjoint folds (typically $k=5$), train a blind diagnostic on the non-visual content of $k{-}1$ folds, and score the held-out fold (\cref{fig:tst_overview:diagnostic_pipeline}).
After $k$ rounds, every question has been scored by a model that never trained on it.
Folds are drawn at random over questions; \cref{tab:rf_tst_results} and the Limitations section also report folds grouped by scene or image.

For each question, the diagnostic's score is the probability it assigns to the gold option for multiple-choice (MC) questions (for TsT-LLM, renormalized over the option letters) and the mean relative accuracy (MRA) of its generated answer for numerical (NUM) questions.
TsT yields two outputs:
\begin{enumerate}[noitemsep,topsep=0pt,parsep=0pt,partopsep=0pt,leftmargin=1.5em]
    \item \textbf{TsT score:}
        the mean held-out score over all questions, an estimate of the benchmark's non-visual solvability.
        It is a lower bound on what a stronger diagnostic could exploit with the same access.
        It is also a designer-side stress test, not ordinary model performance: the diagnostic trains on gold answers from the other folds, which a model under evaluation never sees.
    \item \textbf{Bias score $s(x)$:}
        the held-out score of question $x$.
        A high $s(x)$ flags $x$ as likely answerable without visual input and a candidate for pruning (\cref{sec:mitigating_shortcuts}).
\end{enumerate}

\subsection{LLM-Based TsT Diagnostic}\label{subsec:llm_tst}

Our primary diagnostic fine-tunes a language model (Qwen2-7B;~\citealp{yang2024qwen2}) with LoRA~\citep{hu2021lora} on the question and answer-option text of the training folds and scores the held-out fold.
It needs no feature engineering.
Comparing its TsT score with the same model's zero-shot (ZS) score separates pretrained knowledge from what it learns from the test set.
$\Delta_{\text{TsT}}$ is the TsT score minus the ZS score, and $\Delta s(x)=s(x)-s_{\mathrm{ZS}}(x)$ is its per-question counterpart.
Ranking by $s(x)$ also surfaces questions answerable from pretrained knowledge; ranking by $\Delta s(x)$ targets what is learnable from the test set itself.
The full $k{=}5$ procedure takes about 36 minutes for VSI-Bench (5,130 questions) and 14 minutes for MMMU on one H200 GPU (\cref{app:llm_tst_details}).

\subsection{RF-Based TsT Diagnostic}\label{subsec:rf_tst}

TsT-RF, a secondary diagnostic for template-based benchmarks, is a random forest~\citep{breiman2001random} on hand-crafted non-visual features such as question structure, answer properties, and task-specific statistics (\cref{app:rf_features}).
Its benchmark-level MC score is top-1 accuracy; it has no zero-shot counterpart, so it ranks questions by $s(x)$.
TsT-RF trades generality for interpretability and efficiency: it trains in seconds on a CPU, and its feature importances show which cues drive it.
On VSI-Bench object size estimation, the object category's mean log size alone has Gini importance 0.968: categories such as dishwashers barely vary in size, so designers can drop their questions or sample more varied instances (\cref{app:rf_interpretability}).
Most of its features are \emph{label-conditioned}: statistics of gold answers in the training folds, such as a category's mean size; none of those we use is keyed on the held-out question's own answer. Its inputs, not only its targets, carry gold-answer information that a model under evaluation never sees (\cref{app:rf_leakage}), so we treat TsT-RF scores as designer-side audit statistics, not as blind-model performance.

\begin{table}[t]
    \centering
    \caption{
        \textbf{Learnable non-visual shortcuts are large on template-based benchmarks and small elsewhere.}
        A blind Qwen2-7B is scored zero-shot (ZS) and after $k$-fold LoRA fine-tuning on the other folds of the test set (TsT), with the same rule (\cref{app:llm_tst_details}): mean gold-option probability for MC and MRA for NUM questions, averaged over questions; VSI-Bench (MC+NUM) weights the two formats by question count.
        $\Delta_{\text{TsT}}$ is TsT minus ZS, computed from unrounded scores.
        GPT-4o blind top-1 accuracy is shown for context and is not directly comparable with the Qwen columns.
    }\label{tab:llm_tst_results}
    \small
    \begin{tabular}{lccc@{\hspace{1.5em}}c}
        \toprule
        & \multicolumn{3}{c}{\textbf{Qwen2-7B, blind}} & \\
        \cmidrule(lr){2-4}
        Benchmark & ZS & TsT & $\Delta_{\text{TsT}}$ & \shortstack{GPT-4o, blind\\top-1 acc.} \\
        \midrule
        CV-Bench           & 47.3 & 60.3 & +13.1$^{*}$ & 44.8 \\
        VSI-Bench (MC)     & 25.5 & 39.1 & +13.6 & 34.0 \\
        VSI-Bench (NUM)    & 24.0 & 46.0 & +21.9 & -- \\
        VSI-Bench (MC+NUM) & 24.7 & 42.7 & +17.9 & -- \\
        \midrule
        VideoMME           & 29.2 & 33.4 & +4.1  & 46.6 \\
        MMMU (val, MC)     & 34.2 & 33.6 & $-$0.7 & 52.3 \\
        \midrule
        MMStar (control)   & 24.2 & 29.5 & +5.4  & -- \\
        \bottomrule
        \multicolumn{5}{l}{\scriptsize $^{*}$Scoring zero-shot non-answers as wrong gives +18.4 (\cref{app:llm_tst_details}).}
    \end{tabular}
\end{table}

\subsection{Empirical Validation: Shortcut Susceptibility Varies Across Benchmarks}\label{subsec:diagnostics_validation}

\noindent\textbf{TsT-LLM Results.}
\cref{tab:llm_tst_results} compares a blind Qwen2-7B's ZS score with its TsT score on VSI-Bench, CV-Bench, VideoMME~\citep{fu2024mmecomprehensiveevaluationbenchmark}, and MMMU, with one configuration for every benchmark (\cref{app:llm_tst_details}).
On the template-based VSI-Bench and CV-Bench, TsT-LLM gains 17.9 and 13.1 points over zero-shot, so much of each benchmark can be answered by learning patterns in its non-visual text alone.
VideoMME gains little (+4.1).
On the knowledge-intensive, non-templated MMMU, fine-tuning changes the score little ($\Delta_{\text{TsT}} = {-0.7}$ by gold-option probability; +5.4 by top-1 accuracy, about the gain on MMStar below). GPT-4o~\citep{hurst2024gpto}, however, correctly answers 52.3\% of its MC questions without the image.\footnote{GPT-4o (\texttt{gpt-4o-2024-08-06}, temperature 0, at most 10 output tokens, letter-labeled options) is scored by top-1 accuracy on MC questions (all 2,490 on VSI-Bench; 847 on MMMU); responses with no parseable option, mostly refusals, count as incorrect.}
Blind success on MMMU therefore points mostly to pretrained knowledge, not to patterns learnable from the test set.

\noindent\textbf{Vision-indispensable control.}
On MMStar~\citep{chen2024mmstar}, curated to require vision, Qwen2-7B's blind zero-shot score (24.2) is close to the 25\% uniform baseline and to its score on VSI-Bench's MC questions (25.5). Yet TsT adds only 5.4 points on MMStar, against 13.6 on VSI-Bench's MC questions.
TsT reports no large learnable shortcuts where few are expected, and separates two benchmarks that a blind zero-shot test scores alike.

\noindent\textbf{TsT-RF Results.}
TsT-RF reaches 43.0\% on VSI-Bench, against its 34.0\% per-task frequency baseline, and 56.1\% on CV-Bench, against 41.8\% uniform chance (\cref{tab:rf_tst_results}, which also reports grouped folds). At the benchmark level it points the same way as TsT-LLM.

\section{Mitigating Non-Visual Shortcuts Guided by TsT Insights}\label{sec:mitigating_shortcuts}

We evaluate pruning by its effect on an MLLM's blind score and on its vision-blind gap, its score with vision minus its score without. We compare every pruning method with random removal at the same rate.

\subsection{Iterative Bias Pruning (IBP)}\label{subsec:ibp_methodology}

IBP removes the questions with the highest $s(x)$ in batches and re-diagnoses the rest after each batch (\cref{alg:ibp}), following AFLite~\citep{lebras2020adversarial}, whose ablation found iterative filtering more effective than a single pass.
We re-diagnose since removing questions changes which of the remaining ones are predictable and can create new artifacts~\citep{gururangan2018annotation}.
A removal budget $B$ trades bias reduction against coverage, the batch size $b$ sets how often to re-diagnose, and a threshold $\tau$ on the largest remaining score stops the procedure early.
\Cref{alg:ibp} accepts any per-question score (\cref{app:ibp_details}); our IBP runs rank by TsT-RF $s(x)$ on VSI-Bench (\cref{tab:vsi_provenance}) and by TsT-LLM $\Delta s(x)$ on MMMU (\cref{tab:mmmu_negative_control}).

\begin{algorithm}[h]
    \caption{\methodDebiasLong{} (IBP)}\label{alg:ibp}
    \KwIn{Dataset $\mathcal{D}$; bias-diagnosis function $\mathrm{ComputeSampleBiasScores}(\cdot)$; removal budget $B$;
        batch size $b$; early-stopping threshold $\tau$}
    \KwOut{Debiased dataset $\mathcal{D}'$}

    $R \leftarrow 0$ \tcp*[r]{samples removed so far}
    \While{$R < B$}{
        $\{\,s_i\}_{x_i \in \mathcal{D}} \leftarrow \mathrm{ComputeSampleBiasScores}(\mathcal{D})$ \tcp*[r]{re-diagnose}
        \If{$ \max_{i} s_i \le \tau$}{
            \textbf{break} \tcp*[r]{early stop:\ all biases below threshold}
        }
        $m \leftarrow \min\bigl(b,\; B - R\bigr)$\;
        $\mathcal{I} \leftarrow \mathrm{SelectBatch}\bigl(\{\,s_i\}_{x_i \in \mathcal{D}}, m, \mathcal{D}\bigr)$ \tcp*[r]{select batch for removal}
        $\mathcal{D} \leftarrow \mathcal{D} \setminus \mathcal{I}$\;
        $R \leftarrow R + |\mathcal{I}|$\;
    }
    \Return{$\mathcal{D}' \leftarrow \mathcal{D}$}
\end{algorithm}

\subsection{Pruning VSI-Bench}\label{subsec:mitigation_vsi_robust}

\begin{table}[tb]
    \centering
    \caption{
        \textbf{VSI-Bench-Debiased (designer-in-the-loop pilot).}
        LLaVA-Video-7B scores (MC accuracy and NUM MRA, averaged over questions) with vision (Vis.) and without (Blind), before and after fine-tuning on VSI-Train-10k; Gap is Vis.\ minus Blind.
        On the original set, fine-tuning raises vision and blind scores almost equally (+20.4 vs.\ +18.8); on the debiased set it raises vision much more than blind (+17.4 vs.\ +11.7), so more of the gain there requires vision.
    }\label{tab:vsi_robust_results}
    \small
    \begin{tabular}{lcccccc}
        \toprule
        & \multicolumn{3}{c}{\textbf{VSI-Bench (Original)}} &
        \multicolumn{3}{c}{\textbf{VSI-Bench-Debiased}}\\
        \cmidrule(lr){2-4}\cmidrule(lr){5-7}
        Model & Vis.\ & Blind & Gap & Vis.\ & Blind & Gap \\
        \midrule
        LLaVA-Video-7B (base)    & 36.7 & 25.9 & \textit{10.8} & 31.3 & 20.3 & \textit{11.0} \\
        \quad + VSI-Train-10k FT & 57.1 & 44.7 & \textit{12.4} & 48.7 & 32.0 & \textit{16.7} \\
        \rowcolor{gray!5}
        \color{inctext}\quad\textit{Gain from FT}
            & \color{inctext}\textit{+20.4} & \color{inctext}\textit{+18.8} & \color{inctext}\textit{+1.6}
            & \color{inctext}\textit{+17.4} & \color{inctext}\textit{+11.7} & \color{inctext}\textit{+5.7} \\
        \bottomrule
    \end{tabular}%
\end{table}

VSI-Bench-Debiased is a designer-in-the-loop pilot, not the output of \cref{alg:ibp}. Hand-written filters for each question type remove 2,768 of 5,130 questions (54.0\%). They use only answer statistics, such as how close an object's size is to its category mean, the cue that dominates TsT-RF on object size (\cref{subsec:rf_tst}), or how often an option or object pair is the answer.

The pilot preferentially removes blind-exploitable questions, so the drop in blind score is the primary signal. For LLaVA-Video-7B fine-tuned on VSI-Train-10k, the blind score falls from 44.7 to 32.0, while the vision score falls less, from 57.1 to 48.7 (\cref{tab:vsi_robust_results}). The vision-blind gap thus widens from 12.4 to 16.7, whereas removing 54\% of questions at random leaves it at 12.4 on average (\cref{tab:vsi_random_control}). For Cambrian-S~\citep{yang2025cambrian-s}, a video MLLM for spatial reasoning, the pilot widens the gap from 19.2 to 26.0, against 19.4 for random removal. The widening therefore depends on which questions are removed. Its size varies by model: it is large only for these two, which have the highest original blind scores (\cref{tab:vsi_transfer}), so we treat it as secondary.

Automated per-type IBP on TsT-RF scores widens the gap less than the pilot but more than random removal (\cref{tab:vsi_provenance}). It runs within each question type, with uniform per-type budgets (the same fraction from every type) or the pilot's per-type budgets (its removal count for each type). At the pilot's 54\% removal rate, per-type IBP reaches gaps of 14.1 to 14.7 across the two budget schemes, against 12.2 to 12.4 for random removal of the same number of questions, and 16.7 for the pilot. Per-format IBP, which ranks MC and NUM questions separately, removes 19.5\% of questions at $B{=}1000$ and leaves the gap at 12.4, close to random removal at that rate (12.2).

VSI-Bench-Debiased keeps only 2,362 questions and changes the question-type mix: it removes 85.7\% of medium but only 2.3\% of easy relative-direction questions (\cref{tab:vsi_debiased_statistics}). We therefore release it as a separate evaluation artifact, not a drop-in replacement for VSI-Bench.

\subsection{Pruning MMMU}\label{subsec:ibp_mmmu}

We apply IBP with TsT-LLM to the 900-question MMMU validation set~\citep{yue2023mmmu} as a negative control, since TsT finds few learnable shortcuts there (\cref{subsec:diagnostics_validation}). Ranking its 847 MC questions by $\Delta s(x)$, we remove 100 (11.1\% of the set) and keep all 53 open-ended questions, which TsT-LLM does not score.

On the remaining 800 questions, LLaVA-OneVision-7B's gap moves only from 6.7 to 7.5, inside the range of 1,000 matched random removals (4.62 to 8.75; \cref{tab:mmmu_negative_control}). The removed questions are mostly general-knowledge questions, consistent with MMMU's near-zero $\Delta_{\text{TsT}}$. The control is weak, however: the same model's gap also barely changes on VSI-Bench-Debiased ($-0.5$; \cref{tab:vsi_transfer}).

\section{Related Work}\label{sec:related_work}

\noindent\textbf{Partial-input and blind tests.}
Scoring a benchmark from part of its input is a long-standing check.
VQA reported question-only baselines~\citep{antol2015vqa}, VQA~v2 was compared against a language-only model~\citep{goyal2017making}, and on multiple-choice Visual7W, models given only the question and candidate answers were competitive with models that also saw the image~\citep{jabri2016revisiting}.
In NLI, hypothesis-only classifiers predict the label well above the majority class~\citep{gururangan2018annotation,poliak2018hypothesis}.
For MLLMs, the same check is the ``blind'' test, which MMStar and Cambrian-1 ran across many benchmarks~\citep{chen2024mmstar,tong2024cambrian}.
Related diagnostics score per-question modality importance~\citep{park2024modality}, reliance on local versus global image content~\citep{agarwal2024rci}, and unimodal bias~\citep{chen2024quantifying}.
TsT is also a partial-input probe, with the image as the omitted input, but it measures something different.
A blind test asks what a prompted model answers without the image; TsT asks how much more a text-only model answers after training on the other folds of the same test set.
The two can disagree: blind zero-shot scores on VSI-Bench's MC questions and on MMStar, which was already filtered with blind LLMs, are similar, but their TsT gains differ widely (\cref{subsec:diagnostics_validation}).

\noindent\textbf{Filtering benchmark items.}
These probes have also been used to revise benchmarks.
\citet{gururangan2018annotation} used a hypothesis-only classifier, trained on the training split, to split NLI test sets into Easy and Hard subsets.
MLLM benchmarks use the blind test itself as a filter. MMStar excluded candidates that LLMs answered correctly without the image, then chose its final set by manual review~\citep{chen2024mmstar}; MMMU-Pro begins by removing MMMU questions that text-only LLMs answer correctly~\citep{yue2024mmmupro}.
Adversarial filtering in SWAG and HellaSwag replaces machine-generated distractors that discriminators reject easily~\citep{zellers2018swag,zellers2019hellaswag}.
AFLite~\citep{lebras2020adversarial}, used to build WinoGrande~\citep{sakaguchi2020winogrande}, instead removes items: it repeatedly drops those that classifiers trained on other partitions predict most reliably, and compares the result with size-matched random subsets.
IBP (\cref{alg:ibp}) follows AFLite's loop and its random-removal control.
Partial-input probes, cross-validated per-item scores, and iterative filtering all predate this work; what differs is the probe.
AFLite's classifiers usually see the full input, whereas ours sees only the non-visual input and is fit within the released test set.
TsT-LLM also subtracts the same model's zero-shot score, so $\Delta_{\text{TsT}}$ separates what a pretrained model already knows from what it learns from the test set.

\noindent\textbf{Debiasing VQA.}
Some VQA work changed the evaluation data. VQA~v2 pairs questions with complementary images that lead to different answers~\citep{goyal2017making}, and VQA-CP re-splits VQA so that answer priors differ between training and test~\citep{agrawal2018don}. Later work mines counterexample subsets~\citep{dancette2021beyond} or synthesizes evaluation data to remove answer-option biases~\citep{wang2023dataset}.
Most work instead takes the benchmark as given and changes the model. Approaches include grounded architectures~\citep{agrawal2018don}, adversarial regularization against a question-only branch~\citep{ramakrishnan2018overcoming}, re-weighting examples that a question-only branch answers~\citep{cadene2019rubi,liang2021lpf}, counterfactual inference or training~\citep{niu2021counterfactual,abbasnejad2020counterfactual}, and generative bias models~\citep{cho2023generative}. \citet{belinkov2019adversarial} remove hypothesis-only bias from NLI models with adversarial training.
TsT is dataset-side work.
Its probe resembles these question-only branches, but it is fit on the released test set, and its scores revise the benchmark rather than train a less biased model.

\noindent\textbf{Robust benchmarks and shortcut learning.}
Winoground~\citep{thrush2022winoground}, Vinoground~\citep{zhang2024vinoground}, and the Hateful Memes Challenge~\citep{kiela2020hateful} were designed from the start to resist shortcuts, and a broader literature studies why models prefer available but non-causal features~\citep{geirhos2020shortcut,hermann2023foundations,du2021towards}.
Visual shortcuts matter too; we focus on non-visual ones because they violate the stated intent of a vision-centric benchmark.

\section*{Limitations}\label{sec:limitations}

\noindent\textbf{Fold construction.}
Folds split questions, not the scenes or images behind them, so random folds also credit regularities shared by questions about one scene. Grouping VSI-Bench folds by scene lowers TsT-RF from 43.0 to 38.8 (\cref{tab:rf_tst_results}), mostly in appearance order; object-size and counting priors remain; the TsT-LLM gain changes little (17.9 to 17.0 points). Designers should report both random- and grouped-fold scores.

\noindent\textbf{Remaining shortcuts.}
VSI-Bench-Debiased reduces non-visual shortcuts without removing them, and pruning can create new answer priors. We did not re-run TsT on the pruned set; any refined version should be re-diagnosed with grouped folds before release.

\noindent\textbf{Null results.}
A small $\Delta_{\text{TsT}}$, as on MMMU, is only a lower bound on learnable non-visual signal. A stronger diagnostic may find more, and no partial-input probe detects artifacts involving the omitted image~\citep{feng2019misleading}.

\noindent\textbf{Dependence on the diagnostic.}
TsT scores depend on the diagnostic's configuration: on CV-Bench, the scoring rule and rerun variation each move $\Delta_{\text{TsT}}$ by several points (\cref{app:llm_tst_details}). TsT-LLM and TsT-RF select largely different questions for removal (\cref{tab:cross_diagnostician}), and rankings change substantially across base models (\cref{app:diag_dependence}). TsT-RF needs hand-crafted features for each benchmark, and we have not reviewed flagged questions systematically; \cref{app:rf_interpretability} shows examples.

\section{Conclusion}\label{sec:conclusion}

Models can score well on multimodal benchmarks without using the image, so benchmark scores can overstate visual understanding.
We argue that benchmark designers should ``train on the test set'' before release.
TsT cross-validates a blind model on the test set's text, yielding a benchmark-level TsT score and a per-question bias score $s(x)$.
Learnable shortcuts vary widely. On the template-based VSI-Bench and CV-Bench, TsT-LLM lifts a blind Qwen2-7B 17.9 and 13.1 points above its zero-shot score; on knowledge-intensive MMMU it learns little, although GPT-4o correctly answers 52.3\% of MMMU's multiple-choice questions without the image.
On VSI-Bench, our designer-in-the-loop pilot, released as VSI-Bench-Debiased, lowers a fine-tuned model's blind score from 44.7 to 32.0 and widens its vision-blind gap more than random removal at the same rate. Automated per-type IBP widens it less, though still more than random removal. On MMMU, pruning by $\Delta s(x)$ behaves like random removal.

Open directions include other modalities, open-ended answers, transfer across diagnostics, and rewriting questions or rebalancing answers instead of removing them.
We release VSI-Bench-Debiased on the project page and our code at \url{https://github.com/vision-x-nyu/test-set-training}.
Test-set stress-testing should become a standard step in multimodal benchmark design.

\section*{Acknowledgments}
We thank Shengbang Tong and Anjali Gupta for reviewing this manuscript, Chris Hoang for helpful discussions, and Zaid Khan for advice on LoRA frameworks.
E.B. is supported by the DoD NDSEG Fellowship Program, and S.X. by the MSIT IITP grant (RS-2024-00457882) and NSF award IIS-2443404.

\clearpage
\bibliography{references}

\begin{thebibliography}{50}
\providecommand{\natexlab}[1]{#1}
\providecommand{\url}[1]{\texttt{#1}}
\expandafter\ifx\csname urlstyle\endcsname\relax
  \providecommand{\doi}[1]{doi: #1}\else
  \providecommand{\doi}{doi: \begingroup \urlstyle{rm}\Url}\fi

\bibitem[Abbasnejad et~al.(2020)Abbasnejad, Teney, Parvaneh, Shi, and van~den
  Hengel]{abbasnejad2020counterfactual}
Ehsan Abbasnejad, Damien Teney, Amin Parvaneh, Javen Shi, and Anton van~den
  Hengel.
\newblock Counterfactual vision and language learning.
\newblock In \emph{CVPR}, 2020.

\bibitem[Agarwal et~al.(2025)Agarwal, Patel, Panda, Meghwani, Singh, Dua, Li,
  Sheng, Ravi, and Roth]{agarwal2024rci}
Amit Agarwal, Hitesh~Laxmichand Patel, Srikant Panda, Hansa Meghwani, Jyotika
  Singh, Karan Dua, Paul Li, Tao Sheng, Sujith Ravi, and Dan Roth.
\newblock {RCI}: A score for evaluating global and local reasoning in
  multimodal benchmarks.
\newblock In \emph{EMNLP (Industry Track)}, pages 138--157, 2025.
\newblock \doi{10.18653/v1/2025.emnlp-industry.10}.

\bibitem[Agrawal et~al.(2018)Agrawal, Batra, Parikh, and
  Kembhavi]{agrawal2018don}
Aishwarya Agrawal, Dhruv Batra, Devi Parikh, and Aniruddha Kembhavi.
\newblock Don't just assume; look and answer: Overcoming priors for visual
  question answering.
\newblock In \emph{CVPR}, 2018.

\bibitem[Antol et~al.(2015)Antol, Agrawal, Lu, Mitchell, Batra, Zitnick, and
  Parikh]{antol2015vqa}
Stanislaw Antol, Aishwarya Agrawal, Jiasen Lu, Margaret Mitchell, Dhruv Batra,
  C~Lawrence Zitnick, and Devi Parikh.
\newblock {VQA: Visual Question Answering}.
\newblock In \emph{ICCV}, 2015.

\bibitem[Baruch et~al.(2021)Baruch, Chen, Dehghan, Dimry, Feigin, Fu, Gebauer,
  Joffe, Kurz, Schwartz, and Shulman]{dehghan2021arkitscenes}
Gilad Baruch, Zhuoyuan Chen, Afshin Dehghan, Tal Dimry, Yuri Feigin, Peter Fu,
  Thomas Gebauer, Brandon Joffe, Daniel Kurz, Arik Schwartz, and Elad Shulman.
\newblock {ARK}itscenes - a diverse real-world dataset for 3d indoor scene
  understanding using mobile {RGB}-d data.
\newblock In \emph{NeurIPS}, 2021.

\bibitem[Belinkov et~al.(2019)Belinkov, Poliak, Shieber, Van~Durme, and
  Rush]{belinkov2019adversarial}
Yonatan Belinkov, Adam Poliak, Stuart Shieber, Benjamin Van~Durme, and
  Alexander Rush.
\newblock On adversarial removal of hypothesis-only bias in natural language
  inference.
\newblock In \emph{*SEM}, pages 256--262. Association for Computational
  Linguistics, 2019.
\newblock \doi{10.18653/v1/S19-1028}.
\newblock URL \url{https://aclanthology.org/S19-1028/}.

\bibitem[Breiman(2001)]{breiman2001random}
Leo Breiman.
\newblock Random forests.
\newblock \emph{Machine learning}, 2001.

\bibitem[Cadene et~al.(2019)Cadene, Dancette, Ben-younes, Cord, and
  Parikh]{cadene2019rubi}
Remi Cadene, Corentin Dancette, Hedi Ben-younes, Matthieu Cord, and Devi
  Parikh.
\newblock {RUBi}: Reducing unimodal biases for visual question answering.
\newblock \emph{NeurIPS}, 2019.

\bibitem[Chen et~al.(2024{\natexlab{a}})Chen, Li, Dong, Zhang, Zang, Chen,
  Duan, Wang, Qiao, Lin, and Zhao]{chen2024mmstar}
Lin Chen, Jinsong Li, Xiaoyi Dong, Pan Zhang, Yuhang Zang, Zehui Chen, Haodong
  Duan, Jiaqi Wang, Yu~Qiao, Dahua Lin, and Feng Zhao.
\newblock Are we on the right way for evaluating large vision-language models?
\newblock In \emph{NeurIPS}, pages 27056--27087, 2024{\natexlab{a}}.
\newblock \doi{10.52202/079017-0850}.

\bibitem[Chen et~al.(2024{\natexlab{b}})Chen, Cao, Zhang, and
  Lu]{chen2024quantifying}
Meiqi Chen, Yixin Cao, Yan Zhang, and Chaochao Lu.
\newblock Quantifying and mitigating unimodal biases in multimodal large
  language models: A causal perspective.
\newblock In \emph{Findings of EMNLP}, pages 16449--16469, 2024{\natexlab{b}}.
\newblock \doi{10.18653/v1/2024.findings-emnlp.960}.

\bibitem[Cho et~al.(2023)Cho, Kim, Ryu, and Kweon]{cho2023generative}
Jae~Won Cho, Dong-Jin Kim, Hyeonggon Ryu, and In~So Kweon.
\newblock Generative bias for robust visual question answering.
\newblock In \emph{CVPR}, 2023.

\bibitem[Dai et~al.(2017)Dai, Chang, Savva, Halber, Funkhouser, and
  Nie{\ss}ner]{dai2017scannet}
Angela Dai, Angel~X Chang, Manolis Savva, Maciej Halber, Thomas Funkhouser, and
  Matthias Nie{\ss}ner.
\newblock Scannet: Richly-annotated 3d reconstructions of indoor scenes.
\newblock In \emph{CVPR}, 2017.

\bibitem[Dancette et~al.(2021)Dancette, Cadene, Teney, and
  Cord]{dancette2021beyond}
Corentin Dancette, Remi Cadene, Damien Teney, and Matthieu Cord.
\newblock Beyond question-based biases: Assessing multimodal shortcut learning
  in visual question answering.
\newblock In \emph{ICCV}, pages 1554--1563, 2021.
\newblock \doi{10.1109/ICCV48922.2021.00160}.

\bibitem[Deng et~al.(2009)Deng, Dong, Socher, Li, Li, and
  Fei-Fei]{deng2009imagenet}
Jia Deng, Wei Dong, Richard Socher, Li-Jia Li, Kai Li, and Li~Fei-Fei.
\newblock Imagenet: A large-scale hierarchical image database.
\newblock In \emph{CVPR}, 2009.

\bibitem[Du et~al.(2021)Du, Manjunatha, Jain, Deshpande, Dernoncourt, Gu, Sun,
  and Hu]{du2021towards}
Mengnan Du, Varun Manjunatha, Rajiv Jain, Ruchi Deshpande, Franck Dernoncourt,
  Jiuxiang Gu, Tong Sun, and Xia Hu.
\newblock Towards interpreting and mitigating shortcut learning behavior of
  {NLU} models.
\newblock In \emph{NAACL}, pages 915--929, 2021.
\newblock \doi{10.18653/v1/2021.naacl-main.71}.

\bibitem[Feng et~al.(2019)Feng, Wallace, and Boyd-Graber]{feng2019misleading}
Shi Feng, Eric Wallace, and Jordan Boyd-Graber.
\newblock Misleading failures of partial-input baselines.
\newblock In \emph{ACL}, pages 5533--5538. Association for Computational
  Linguistics, 2019.
\newblock \doi{10.18653/v1/P19-1554}.
\newblock URL \url{https://aclanthology.org/P19-1554/}.

\bibitem[Fu et~al.(2025)Fu, Dai, Luo, Li, Ren, Zhang, Wang, Zhou, Shen, Zhang,
  et~al.]{fu2024mmecomprehensiveevaluationbenchmark}
Chaoyou Fu, Yuhan Dai, Yongdong Luo, Lei Li, Shuhuai Ren, Renrui Zhang, Zihan
  Wang, Chenyu Zhou, Yunhang Shen, Mengdan Zhang, et~al.
\newblock {Video-MME}: The first-ever comprehensive evaluation benchmark of
  multi-modal {LLMs} in video analysis.
\newblock In \emph{CVPR}, 2025.

\bibitem[Geirhos et~al.(2020)Geirhos, Jacobsen, Michaelis, Zemel, Brendel,
  Bethge, and Wichmann]{geirhos2020shortcut}
Robert Geirhos, J{\"o}rn-Henrik Jacobsen, Claudio Michaelis, Richard Zemel,
  Wieland Brendel, Matthias Bethge, and Felix~A Wichmann.
\newblock Shortcut learning in deep neural networks.
\newblock \emph{Nature Machine Intelligence}, 2\penalty0 (11):\penalty0
  665--673, 2020.
\newblock \doi{10.1038/s42256-020-00257-z}.

\bibitem[Goyal et~al.(2017)Goyal, Khot, Summers-Stay, Batra, and
  Parikh]{goyal2017making}
Yash Goyal, Tejas Khot, Douglas Summers-Stay, Dhruv Batra, and Devi Parikh.
\newblock {Making the V in VQA matter: Elevating the role of image
  understanding in visual question answering}.
\newblock In \emph{CVPR}, 2017.

\bibitem[Gururangan et~al.(2018)Gururangan, Swayamdipta, Levy, Schwartz,
  Bowman, and Smith]{gururangan2018annotation}
Suchin Gururangan, Swabha Swayamdipta, Omer Levy, Roy Schwartz, Samuel~R.
  Bowman, and Noah~A. Smith.
\newblock Annotation artifacts in natural language inference data.
\newblock In \emph{NAACL}, pages 107--112. Association for Computational
  Linguistics, 2018.
\newblock \doi{10.18653/v1/N18-2017}.
\newblock URL \url{https://aclanthology.org/N18-2017/}.

\bibitem[Hastie et~al.(2009)Hastie, Tibshirani, and
  Friedman]{hastie2009elements}
Trevor Hastie, Robert Tibshirani, and Jerome~H. Friedman.
\newblock \emph{The elements of statistical learning: data mining, inference,
  and prediction}.
\newblock Springer, 2 edition, 2009.
\newblock \doi{10.1007/978-0-387-84858-7}.

\bibitem[Hermann et~al.(2024)Hermann, Mobahi, Fel, and
  Mozer]{hermann2023foundations}
Katherine~L Hermann, Hossein Mobahi, Thomas Fel, and Michael~C Mozer.
\newblock On the foundations of shortcut learning.
\newblock In \emph{ICLR}, 2024.
\newblock URL \url{https://arxiv.org/abs/2310.16228}.

\bibitem[Hu et~al.(2022)Hu, Shen, Wallis, Allen-Zhu, Li, Wang, Wang, and
  Chen]{hu2021lora}
Edward~J Hu, Yelong Shen, Phillip Wallis, Zeyuan Allen-Zhu, Yuanzhi Li, Shean
  Wang, Lu~Wang, and Weizhu Chen.
\newblock {LoRA: Low-Rank Adaptation of Large Language Models}.
\newblock In \emph{ICLR}, 2022.

\bibitem[Hurst et~al.(2024)Hurst, Lerer, Goucher, Perelman, Ramesh, Clark,
  Ostrow, Welihinda, Hayes, Radford, et~al.]{hurst2024gpto}
Aaron Hurst, Adam Lerer, Adam~P Goucher, Adam Perelman, Aditya Ramesh, Aidan
  Clark, AJ~Ostrow, Akila Welihinda, Alan Hayes, Alec Radford, et~al.
\newblock {GPT-4o} system card.
\newblock \emph{arXiv preprint arXiv:2410.21276}, 2024.

\bibitem[Jabri et~al.(2016)Jabri, Joulin, and van~der
  Maaten]{jabri2016revisiting}
Allan Jabri, Armand Joulin, and Laurens van~der Maaten.
\newblock Revisiting visual question answering baselines.
\newblock In \emph{ECCV}, pages 727--739, 2016.
\newblock \doi{10.1007/978-3-319-46484-8_44}.

\bibitem[Kiela et~al.(2020)Kiela, Firooz, Mohan, Goswami, Singh, Ringshia, and
  Testuggine]{kiela2020hateful}
Douwe Kiela, Hamed Firooz, Aravind Mohan, Vedanuj Goswami, Amanpreet Singh,
  Pratik Ringshia, and Davide Testuggine.
\newblock The hateful memes challenge: Detecting hate speech in multimodal
  memes.
\newblock In \emph{NeurIPS}, 2020.

\bibitem[Le~Bras et~al.(2020)Le~Bras, Swayamdipta, Bhagavatula, Zellers,
  Peters, Sabharwal, and Choi]{lebras2020adversarial}
Ronan Le~Bras, Swabha Swayamdipta, Chandra Bhagavatula, Rowan Zellers, Matthew
  Peters, Ashish Sabharwal, and Yejin Choi.
\newblock Adversarial filters of dataset biases.
\newblock In \emph{ICML}, volume 119 of \emph{Proceedings of Machine Learning
  Research}, pages 1078--1088, 2020.
\newblock URL \url{https://proceedings.mlr.press/v119/bras20a.html}.

\bibitem[LeCun et~al.(1998)LeCun, Bottou, Bengio, and
  Haffner]{lecun1998gradient}
Yann LeCun, Leon Bottou, Yoshua Bengio, and Patrick Haffner.
\newblock Gradient-based learning applied to document recognition.
\newblock \emph{Proceedings of the IEEE}, 1998.

\bibitem[Li et~al.(2025)Li, Zhang, Guo, Zhang, Li, Zhang, Zhang, Zhang, Li,
  Liu, and Li]{li2024llavaov}
Bo~Li, Yuanhan Zhang, Dong Guo, Renrui Zhang, Feng Li, Hao Zhang, Kaichen
  Zhang, Peiyuan Zhang, Yanwei Li, Ziwei Liu, and Chunyuan Li.
\newblock {LLaVA-OneVision}: Easy visual task transfer.
\newblock \emph{Transactions on Machine Learning Research}, 2025.

\bibitem[Liang et~al.(2021)Liang, Hu, and Zhu]{liang2021lpf}
Zujie Liang, Haifeng Hu, and Jiaying Zhu.
\newblock {LPF}: A language-prior feedback objective function for de-biased
  visual question answering.
\newblock In \emph{SIGIR}, 2021.

\bibitem[Lin et~al.(2014)Lin, Maire, Belongie, Hays, Perona, Ramanan,
  Doll{\'a}r, and Zitnick]{lin2014microsoft}
Tsung-Yi Lin, Michael Maire, Serge Belongie, James Hays, Pietro Perona, Deva
  Ramanan, Piotr Doll{\'a}r, and C~Lawrence Zitnick.
\newblock Microsoft coco: Common objects in context.
\newblock In \emph{ECCV}, 2014.

\bibitem[Niu et~al.(2021)Niu, Tang, Zhang, Lu, Hua, and
  Wen]{niu2021counterfactual}
Yulei Niu, Kaihua Tang, Hanwang Zhang, Zhiwu Lu, Xian-Sheng Hua, and Ji-Rong
  Wen.
\newblock Counterfactual {VQA}: A cause-effect look at language bias.
\newblock In \emph{CVPR}, 2021.

\bibitem[Park et~al.(2025)Park, Jang, Alasaly, Mopidevi, Zolensky, Eaton, Lee,
  and Johnson]{park2024modality}
Jean Park, Kuk~Jin Jang, Basam Alasaly, Sriharsha Mopidevi, Andrew Zolensky,
  Eric Eaton, Insup Lee, and Kevin Johnson.
\newblock Assessing modality bias in video question answering benchmarks with
  multimodal large language models.
\newblock In \emph{AAAI}, 2025.

\bibitem[Poliak et~al.(2018)Poliak, Naradowsky, Haldar, Rudinger, and
  Van~Durme]{poliak2018hypothesis}
Adam Poliak, Jason Naradowsky, Aparajita Haldar, Rachel Rudinger, and Benjamin
  Van~Durme.
\newblock Hypothesis only baselines in natural language inference.
\newblock In \emph{*SEM}, pages 180--191. Association for Computational
  Linguistics, 2018.
\newblock \doi{10.18653/v1/S18-2023}.
\newblock URL \url{https://aclanthology.org/S18-2023/}.

\bibitem[Ramakrishnan et~al.(2018)Ramakrishnan, Agrawal, and
  Lee]{ramakrishnan2018overcoming}
Sainandan Ramakrishnan, Aishwarya Agrawal, and Stefan Lee.
\newblock Overcoming language priors in visual question answering with
  adversarial regularization.
\newblock In \emph{NeurIPS}, 2018.

\bibitem[Sakaguchi et~al.(2020)Sakaguchi, Le~Bras, Bhagavatula, and
  Choi]{sakaguchi2020winogrande}
Keisuke Sakaguchi, Ronan Le~Bras, Chandra Bhagavatula, and Yejin Choi.
\newblock {WinoGrande}: An adversarial {Winograd} schema challenge at scale.
\newblock In \emph{AAAI}, pages 8732--8740, 2020.
\newblock \doi{10.1609/aaai.v34i05.6399}.

\bibitem[Stone(1974)]{stone1974cross}
Mervyn Stone.
\newblock Cross-validatory choice and assessment of statistical predictions.
\newblock \emph{Journal of the royal statistical society}, 1974.

\bibitem[Thrush et~al.(2022)Thrush, Jiang, Bartolo, Singh, Williams, Kiela, and
  Ross]{thrush2022winoground}
Tristan Thrush, Ryan Jiang, Max Bartolo, Amanpreet Singh, Adina Williams, Douwe
  Kiela, and Candace Ross.
\newblock Winoground: Probing vision and language models for visio-linguistic
  compositionality.
\newblock In \emph{CVPR}, 2022.

\bibitem[Tong et~al.(2024)Tong, Brown, Wu, Woo, Middepogu, Akula, Yang, Yang,
  Iyer, Pan, Wang, Fergus, LeCun, and Xie]{tong2024cambrian}
Shengbang Tong, Ellis Brown, Penghao Wu, Sanghyun Woo, Manoj Middepogu,
  Sai~Charitha Akula, Jihan Yang, Shusheng Yang, Adithya Iyer, Xichen Pan,
  Austin Wang, Rob Fergus, Yann LeCun, and Saining Xie.
\newblock {Cambrian-1: A Fully Open, Vision-Centric Exploration of Multimodal
  LLMs}.
\newblock \emph{NeurIPS}, 2024.

\bibitem[Wang et~al.(2023)Wang, Chen, You, Xu, He, Li, Codella, Chang, and
  Chang]{wang2023dataset}
Zhecan Wang, Long Chen, Haoxuan You, Keyang Xu, Yicheng He, Wenhao Li, Noel
  Codella, Kai-Wei Chang, and Shih-Fu Chang.
\newblock Dataset bias mitigation in multiple-choice visual question answering
  and beyond.
\newblock In \emph{Findings of EMNLP}, pages 8598--8617. Association for
  Computational Linguistics, 2023.
\newblock \doi{10.18653/v1/2023.findings-emnlp.576}.
\newblock URL \url{https://aclanthology.org/2023.findings-emnlp.576/}.

\bibitem[Yang et~al.(2024)Yang, Yang, Hui, Zheng, Yu, Zhou,
  et~al.]{yang2024qwen2}
An~Yang, Baosong Yang, Binyuan Hui, Bo~Zheng, Bowen Yu, Chang Zhou, et~al.
\newblock Qwen2 technical report.
\newblock \emph{arXiv preprint arXiv:2407.10671}, 2024.
\newblock \doi{10.48550/arXiv.2407.10671}.
\newblock URL \url{https://arxiv.org/abs/2407.10671}.

\bibitem[Yang et~al.(2025)Yang, Yang, Gupta, Han, Fei-Fei, and
  Xie]{yang2024think}
Jihan Yang, Shusheng Yang, Anjali~W. Gupta, Rilyn Han, Li~Fei-Fei, and Saining
  Xie.
\newblock {Thinking in Space: How Multimodal Large Language Models See,
  Remember and Recall Spaces}.
\newblock \emph{CVPR}, 2025.

\bibitem[Yang et~al.(2026)Yang, Yang, Huang, Brown, Yang, Yu, Tong, Zheng, Xu,
  Wang, Lu, Fergus, LeCun, Fei-Fei, and Xie]{yang2025cambrian-s}
Shusheng Yang, Jihan Yang, Pinzhi Huang, Ellis Brown, Zihao Yang, Yue Yu,
  Shengbang Tong, Zihan Zheng, Yifan Xu, Muhan Wang, Daohan Lu, Rob Fergus,
  Yann LeCun, Li~Fei-Fei, and Saining Xie.
\newblock {Cambrian-S: Towards Spatial Supersensing in Video}.
\newblock In \emph{ICLR}, 2026.

\bibitem[Yeshwanth et~al.(2023)Yeshwanth, Liu, Nie{\ss}ner, and
  Dai]{yeshwanth2023scannet++}
Chandan Yeshwanth, Yueh-Cheng Liu, Matthias Nie{\ss}ner, and Angela Dai.
\newblock Scannet++: A high-fidelity dataset of 3d indoor scenes.
\newblock In \emph{ICCV}, 2023.

\bibitem[Yue et~al.(2024)Yue, Ni, Zhang, Zheng, Liu, Zhang, Stevens, Jiang,
  Ren, Sun, Wei, Yu, Yuan, Sun, Yin, Zheng, Yang, Liu, Huang, Sun, Su, and
  Chen]{yue2023mmmu}
Xiang Yue, Yuansheng Ni, Kai Zhang, Tianyu Zheng, Ruoqi Liu, Ge~Zhang, Samuel
  Stevens, Dongfu Jiang, Weiming Ren, Yuxuan Sun, Cong Wei, Botao Yu, Ruibin
  Yuan, Renliang Sun, Ming Yin, Boyuan Zheng, Zhenzhu Yang, Yibo Liu, Wenhao
  Huang, Huan Sun, Yu~Su, and Wenhu Chen.
\newblock Mmmu: A massive multi-discipline multimodal understanding and
  reasoning benchmark for expert agi.
\newblock In \emph{CVPR}, 2024.

\bibitem[Yue et~al.(2025)Yue, Zheng, Ni, Wang, Zhang, Tong, Sun, Yu, Zhang,
  Sun, Su, Chen, and Neubig]{yue2024mmmupro}
Xiang Yue, Tianyu Zheng, Yuansheng Ni, Yubo Wang, Kai Zhang, Shengbang Tong,
  Yuxuan Sun, Botao Yu, Ge~Zhang, Huan Sun, Yu~Su, Wenhu Chen, and Graham
  Neubig.
\newblock {MMMU-Pro}: A more robust multi-discipline multimodal understanding
  benchmark.
\newblock In \emph{ACL}, pages 15134--15186, 2025.
\newblock \doi{10.18653/v1/2025.acl-long.736}.

\bibitem[Zellers et~al.(2018)Zellers, Bisk, Schwartz, and
  Choi]{zellers2018swag}
Rowan Zellers, Yonatan Bisk, Roy Schwartz, and Yejin Choi.
\newblock {SWAG}: A large-scale adversarial dataset for grounded commonsense
  inference.
\newblock In \emph{EMNLP}, pages 93--104. Association for Computational
  Linguistics, 2018.
\newblock \doi{10.18653/v1/D18-1009}.
\newblock URL \url{https://aclanthology.org/D18-1009/}.

\bibitem[Zellers et~al.(2019)Zellers, Holtzman, Bisk, Farhadi, and
  Choi]{zellers2019hellaswag}
Rowan Zellers, Ari Holtzman, Yonatan Bisk, Ali Farhadi, and Yejin Choi.
\newblock {HellaSwag}: Can a machine really finish your sentence?
\newblock In \emph{ACL}, pages 4791--4800. Association for Computational
  Linguistics, 2019.
\newblock \doi{10.18653/v1/P19-1472}.
\newblock URL \url{https://aclanthology.org/P19-1472/}.

\bibitem[Zhang et~al.(2024)Zhang, Cai, and Lee]{zhang2024vinoground}
Jianrui Zhang, Mu~Cai, and Yong~Jae Lee.
\newblock Vinoground: Scrutinizing {LMMs} over dense temporal reasoning with
  short videos.
\newblock \emph{arXiv preprint arXiv:2410.02763}, 2024.

\bibitem[Zhang et~al.(2025)Zhang, Wu, Li, Li, Ma, Liu, and Li]{zhang2024video}
Yuanhan Zhang, Jinming Wu, Wei Li, Bo~Li, Zejun Ma, Ziwei Liu, and Chunyuan Li.
\newblock {LLaVA-Video}: Video instruction tuning with synthetic data.
\newblock \emph{Transactions on Machine Learning Research}, 2025.

\end{thebibliography}
\bibliographystyle{plainnat}
\clearpage

\appendix
\section*{Appendix}\label{app:appendix}

This appendix gives the LLM-use disclosure (\S\ref{app:llm_disclosure}), VSI-Train-10k generation (\S\ref{app:vsi-train-details}), TsT diagnostic details (\S\ref{app:tst_diagnostic_details}), debiasing details (\S\ref{app:debiasing_details}), and the TsT-RF interpretability case study (\S\ref{app:rf_interpretability}).

\section{LLM Disclosure}\label{app:llm_disclosure}

LLMs were used for drafting and editing assistance, bibliographic and consistency checks, coding assistance, and data visualization in preparing this manuscript.
The authors directed the work, verified the reported claims, citations, and numerical results against the sources, and take responsibility for the final text. Research ideas, experimental design, data generation, and evaluation were done and validated by the authors.

\section{VSI-Train-10k Generation}\label{app:vsi-train-details}
VSI-Train-10k is the in-distribution training set used to fine-tune LLaVA-Video-7B in \cref{tab:vsi_robust_results}.
We build it with the VSI-Bench curation pipeline~\citep{yang2024think,yang2025cambrian-s}, applied to the training splits of ScanNet~\citep{dai2017scannet}, ScanNet++~\citep{yeshwanth2023scannet++}, and ARKitScenes~\citep{dehghan2021arkitscenes}.
We extract object counts, bounding boxes, and room sizes, generate questions with VSI-Bench's templates for its seven rule-based question types (all but route planning), and randomly sample 10K questions, with at most 20 per question type per scene.

\section{TsT Diagnostic Details}\label{app:tst_diagnostic_details}

\subsection{TsT-LLM Implementation Details}\label{app:llm_tst_details}
Every TsT-LLM result in \cref{tab:llm_tst_results} uses Qwen2-7B-Instruct~\citep{yang2024qwen2} (revision \texttt{f2826a0}; datasets VSI-Bench \texttt{d7cb1a3}, CV-Bench \texttt{bc284db}, VideoMME \texttt{ead1408}) with one configuration.
Cross-validation: $k{=}5$, seed 42, one repeat.
LoRA~\citep{hu2021lora}: rank $r{=}8$, $\alpha{=}16$, dropout 0.1, all-module targets.
Training: one epoch per fold, train and evaluation batch size 4, gradient accumulation 1, learning rate $2\times10^{-4}$ with cosine decay and 0.1 warmup, FP16, maximum sequence length 1024, a 10\% validation split of the training folds, and the \texttt{gemma} training template.
Inference: raw prompts (no chat template), temperature 0, top-$p$ 1, at most 10 new tokens, and top-100 log probabilities.

Each question is scored only in its held-out fold: an MC question by the gold option's probability, a NUM question by the MRA of its generated answer.
If the gold letter is not among the top-100 log probabilities, we parse the generated answer instead; an answer with no parseable option receives a random option.
This fallback does not fire after fine-tuning and has little effect on VSI-Bench and VideoMME.
On CV-Bench, however, many zero-shot answers begin with a word rather than an option letter and so receive chance credit.
Scoring them as wrong would give a CV-Bench $\Delta_{\text{TsT}}$ of about +18.4 rather than +13.1; we report the smaller value, since the larger one would credit TsT for learning the answer format.

LoRA training on GPUs is not bit-reproducible. Reruns of this configuration give VSI-Bench (MC+NUM) scores of 42.6--42.8 ($\Delta_{\text{TsT}}{\approx}18$); the CV-Bench gain varies more across reruns and fold assignments, from about 13 to 18 points.

We also ran a separate chat-formatted configuration ($r{=}32$, $\alpha{=}64$, two epochs, batch 32, learning rate $5\times10^{-5}$, cosine decay, 0.1 warmup, and the \texttt{qwen} chat template). Its $\Delta s(x)$ rankings agree with a matched $r{=}128$, $\alpha{=}256$, three-epoch run of the same setup (Spearman correlation 0.87; 70.9\% top-1,000 overlap).

\subsection{TsT-RF Features}\label{app:rf_features}
TsT-RF (\cref{subsec:rf_tst}) uses non-visual features tailored to each benchmark's question templates. They fall into four groups: (1) entities and relations parsed from the question text, such as object categories, distances, and directions; (2) answer-space properties, such as answer format and option count; (3) question-type and object metadata; and (4) task-specific frequency and distributional statistics fit on the training folds, such as the average size of an object category for size estimation. For VSI-Bench, regular expressions parse these features from the question text. On CV-Bench, the forest is trained to predict which option is correct.
\Cref{tab:vsi_features,tab:cvb_features} list representative features for each question type; the implementation also derives closely related frequency and ratio features.

The reported implementation uses scikit-learn forests with no maximum depth: 100 trees for classifiers and 200 for regressors; rankings barely change across forests of 200 to 2,000 trees and depth 10, 20, or unlimited (\cref{app:diag_dependence}).

\paragraph{Avoiding label leakage in TsT-RF features.}\label{app:rf_leakage}
A TsT-RF feature must be computable without the evaluated question's own answer. Statistics fit on the training folds, such as how often an object is the answer to a question type, meet this requirement; a lookup keyed on the held-out question's gold answer does not, because it encodes the label. The features marked $^{\ddagger}$ in \cref{tab:vsi_features_continued,tab:cvb_features} are such lookups and are excluded from every TsT-RF score and automated selection we report. A simple check is label invariance: permuting the labels of the held-out fold must leave all of its feature values unchanged.

\subsection{TsT-RF Results}\label{app:rf_results}
\Cref{tab:rf_tst_results} gives TsT-RF scores with random and grouped folds.

\begin{table}[!htbp]
    \centering
    \caption{
        \textbf{TsT-RF cross-validation scores.}
        Features are label-conditioned (\cref{subsec:rf_tst}), so these are designer-side audit statistics.
        Baselines are for scale: uniform chance over each question's options, averaged over questions, and VSI-Bench's per-task frequency baseline~\citep{yang2024think}.
        Scores are averaged over questions (MC accuracy, plus NUM MRA for VSI-Bench).
        Grouped folds keep all questions about one scene (VSI-Bench) or image (CV-Bench) in the same fold.
        Features marked $^{\ddagger}$ in \cref{tab:vsi_features_continued,tab:cvb_features} are excluded.
        Dataset revisions: CV-Bench \texttt{bc284db}, VSI-Bench \texttt{bc96b17}.
    }\label{tab:rf_tst_results}
    \begin{tabular}{lcccc}
        \toprule
        & & & \multicolumn{2}{c}{TsT-RF CV score} \\
        \cmidrule(lr){4-5}
        Benchmark & Uniform chance & Frequency baseline & Random folds & Grouped folds \\
        \midrule
        CV-Bench     & 41.8 & --   & 56.1 & 55.7 \\
        VSI-Bench    & --   & 34.0 & 43.0 & 38.8 \\
        \bottomrule
    \end{tabular}
\end{table}

\begin{table}[!htbp]
    \centering
    \caption{
        \textbf{Representative VSI-Bench TsT-RF features by question type.}
        Features come from the question text and, for MC questions, the answer options; none uses visual input.
        $^{\ddagger}$Keyed on the held-out question's gold answer; excluded from all reported TsT-RF scores and automated selections.
    }\label{tab:vsi_features}\label{tab:vsi_features_continued}
    \scriptsize
    \setlength{\tabcolsep}{3pt}
    \begin{tabular}{p{0.10\linewidth}p{0.05\linewidth}p{0.11\linewidth}p{0.27\linewidth}>{\raggedright\arraybackslash}p{0.37\linewidth}}
        \toprule
        Task & Format & Feature Type & Feature Name & Description \\
        \midrule
        \multirow{6}{*}{\shortstack[l]{Object\\Counting}} 
        & \multirow{6}{*}{NUM}
        & Categorical
        & \texttt{object} & Object category \\
        \cmidrule{3-5}
        & & \multirow{5}{*}{Numerical}
        & \texttt{obj\_count} & Count of this object \\
        & & & \texttt{obj\_val\_log\_mean} & Mean of log gold values \\
        & & & \texttt{obj\_val\_log\_std} & Std dev of log gold values \\
        & & & \texttt{global\_mean\_log} & Global mean (log space) \\
        & & & \texttt{global\_std\_log} & Global std dev (log space) \\
        \midrule
        \multirow{6}{*}{\shortstack[l]{Object Size\\Estimation}} 
        & \multirow{6}{*}{NUM}
        & Categorical
        & \texttt{object} & Object category \\
        \cmidrule{3-5}
        & & \multirow{5}{*}{Numerical}
        & \texttt{obj\_count} & Count of this object \\
        & & & \texttt{obj\_val\_log\_mean} & Mean of log gold sizes \\
        & & & \texttt{obj\_val\_log\_std} & Std dev of log gold sizes \\
        & & & \texttt{obj\_freq\_score} & Min-max scaled object count \\
        & & & \texttt{obj\_val\_log\_ratio} & One minus scaled std/mean of log sizes \\
        \midrule
        \multirow{4}{*}{\shortstack[l]{Object Abs.\\Distance}} 
        & \multirow{4}{*}{NUM}
        & Categorical &
        \texttt{object\_pair} & Sorted pair of objects \\
        \cmidrule{3-5}
        & & \multirow{3}{*}{Numerical}
        & \texttt{pair\_count} & Count of this pair \\
        & & & \texttt{pair\_val\_mean\_log} & Mean of log distances \\
        & & & \texttt{pair\_val\_std\_log} & Std dev of log distances \\
        \midrule
        \multirow{2}{*}{\shortstack[l]{Room Size\\Estimation}} 
        & \multirow{2}{*}{NUM}
        & \multirow{2}{*}{Numerical}
        & \texttt{global\_mean\_log} & Global mean (log space) \\
        & & & \texttt{global\_std\_log} & Global std dev (log space) \\
        \midrule
        \multirow{8}{*}{\shortstack[l]{Object Rel.\\Distance}} 
        & \multirow{8}{*}{MC}
        & \multirow{2}{*}{Categorical} &
        \texttt{target\_object} & Category of target object \\
        & & & \texttt{object\_\{i\}} & Object category of option $i$, $\forall i \in [4]$ \\
        \cmidrule{3-5}
        & & \multirow{6}{*}{Numerical}
        & \texttt{opt\_\{i\}\_obj\_freq} & Frequency of option $i$'s object being correct \\
        & & & \texttt{opt\_\{i\}\_pair\_freq}$^{\ddagger}$ & Frequency of the (target, gold object) pair being correct\\
        & & & \texttt{opt\_\{i\}\_ord\_pair\_freq}$^{\ddagger}$ & Ordered-pair version of the above\\
        & & & \texttt{max\_opt\_obj\_freq} & Max option frequency \\
        & & & \texttt{max\_opt\_pair\_freq}$^{\ddagger}$ & Max of the pair frequencies \\
        & & & \texttt{max\_opt\_ord\_pair\_freq}$^{\ddagger}$ & Max of the ordered-pair frequencies \\
        \midrule
        \multirow{5}{*}{\shortstack[l]{Object Rel.\\Direction}} 
        & \multirow{5}{*}{MC}
        & \multirow{4}{*}{Categorical}
        & \texttt{difficulty} & Subtype (easy/med/hard) \\
        & & & \texttt{positioning\_object} & Object the viewer stands by \\
        & & & \texttt{orienting\_object} & Object the viewer faces \\
        & & & \texttt{querying\_object} & Query target object \\
        \cmidrule{3-5}
        & & Numerical &
        \texttt{obj\_freq\_score} & Combined frequency of all 3 objects being in a question \\
        \midrule
        \multirow{8}{*}{\shortstack[l]{Route\\Planning}} 
        & \multirow{8}{*}{MC}
        & \multirow{4}{*}{Categorical}
        & \texttt{beginning\_object} & Starting position object \\
        & & & \texttt{facing\_object} & Initial orientation object \\
        & & & \texttt{target\_object} & Destination object \\
        & & & \texttt{opt\_\{i\}} & Route string option $i$ \\
        \cmidrule{3-5}
        & & \multirow{4}{*}{Numerical}
        & \texttt{num\_steps} & Number of steps per route option (equal across options) \\
        & & & \texttt{num\_choices} & Number of MC options \\
        & & & \texttt{opt\_\{i\}\_freq\_score} & Route option $i$ frequency \\
        & & & \texttt{obj\_freq\_score} & Combined frequency of all 3 objects being in a question \\
        \midrule
        \multirow{5}{*}{\shortstack[l]{Appearance\\Order}} 
        & \multirow{5}{*}{MC}
        & Categorical
        & \texttt{opt\_seq\_\{i\}} & Object sequence for option $i$: $[o^i_1, o^i_2, \ldots, o^i_n]$ \\
        \cmidrule{3-5}
        & & \multirow{4}{*}{Numerical}
        & \texttt{seq\_\{i\}\_pos\_score} & Sum of frequencies of object $o^i_j$ appearing in position $j$ in a gold sequence, $\forall j \in [n]$ \\
        & & & \texttt{seq\_\{i\}\_adj\_pair\_score} & Sum of frequencies of adjacent pairs $o^i_j, o^i_{j+1}$ in a gold sequence, $\forall j \in [n-1]$ \\
        & & & \texttt{seq\_\{i\}\_comb\_pair\_score} & Sum of frequencies of all combinatorial pairs $o^i_j, o^i_k$ in a gold sequence, $\forall j, k \in [n],\ j < k$ \\
        & & & \texttt{seq\_\{i\}\_score} & Average of the above three scores for option $i$ \\
        \bottomrule
    \end{tabular}
\end{table}

\begin{table}[!htbp]
    \centering
    \caption{\textbf{Representative CV-Bench TsT-RF features by question type.}
    All tasks are MC; no feature uses visual input.
    $^{\ddagger}$Keyed on the held-out question's gold answer; excluded from all reported TsT-RF scores.
    }\label{tab:cvb_features}
    \scriptsize
    \setlength{\tabcolsep}{4pt}
    \begin{tabular}{p{0.10\linewidth}p{0.11\linewidth}p{0.30\linewidth}>{\raggedright\arraybackslash}p{0.39\linewidth}}
        \toprule
        Task & Feature Type & Feature Name & Description \\
        \midrule
        \multirow{9}{*}{\shortstack[l]{2D\\Count}} 
        & Categorical
        & \texttt{object} & Object category \\
        \cmidrule{2-4}
        & \multirow{8}{*}{Numerical}
        & \texttt{n\_options} & Number of MC choices \\
        & & \texttt{obj\_count} & Count of this object \\
        & & \texttt{obj\_val\_log\_mean} & Mean of log gold values \\
        & & \texttt{obj\_val\_log\_std} & Std dev of log gold values \\
        & & \texttt{global\_mean\_log} & Global mean (log space) \\
        & & \texttt{global\_std\_log} & Global std dev (log space) \\
        & & \texttt{opt\_\{i\}\_dist\_from\_obj\_mean} & Log-space distance of option $i$ from the object's mean count \\
        & & \texttt{opt\_\{i\}\_dist\_from\_global\_mean} & Log-space distance of option $i$ from the global mean count \\
        \midrule
        \multirow{9}{*}{\shortstack[l]{2D Spatial\\Relation}} 
        & Categorical
        & \texttt{object\_\{i\}} & Category of object $i$, $\forall i \in [2]$ \\
        \cmidrule{2-4}
        & \multirow{8}{*}{Numerical}
        & \texttt{n\_options} & Number of MC choices \\
        & & \texttt{pair\_freq\_score} & Frequency of object pair in questions \\
        & & \texttt{pair\_answer\_freq\_score}$^{\ddagger}$ & Training-fold frequency of the gold answer for this pair \\
        & & \texttt{is\_majority\_answer}$^{\ddagger}$ & Whether the gold answer is the training-fold majority \\
        & & \texttt{question\_length} & Question character length \\
        & & \texttt{contains\_\{dir\}} & Contains spatial keyword \texttt{dir}, for \texttt{dir} $\in$ \texttt{\{left, right, above, below, front, behind\}} \\
        & & \texttt{spatial\_keyword\_count} & Total spatial direction keywords \\
        \midrule
        \multirow{4}{*}{\shortstack[l]{3D Depth}} 
        & Categorical
        & \texttt{object\_\{i\}} & Category of object $i$, $\forall i \in [2]$ \\
        \cmidrule{2-4}
        & \multirow{3}{*}{Numerical}
        & \texttt{n\_options} & Number of MC choices \\
        & & \texttt{pair\_freq\_score} & Frequency of object pair in questions \\
        & & \texttt{pair\_answer\_freq\_score}$^{\ddagger}$ & Training-fold frequency of the gold answer for this pair \\
        \midrule
        \multirow{4}{*}{\shortstack[l]{3D Distance}} 
        & Categorical
        & \texttt{object\_\{i\}} & Category of object $i$, $\forall i \in [2]$ \\
        \cmidrule{2-4}
        & \multirow{3}{*}{Numerical}
        & \texttt{n\_options} & Number of MC choices \\
        & & \texttt{pair\_freq\_score} & Frequency of object pair in questions \\
        & & \texttt{pair\_answer\_freq\_score}$^{\ddagger}$ & Training-fold frequency of the gold answer for this pair \\
        \bottomrule
    \end{tabular}
\end{table}

\section{Debiasing Details}\label{app:debiasing_details}

\subsection{IBP Details}\label{app:ibp_details}

In \cref{alg:ibp}, $\mathrm{SelectBatch}(\cdot)$ can remove the highest-scoring questions directly or balance categorical groups, always ranking by the chosen per-question score ($s(x)$ or $\Delta s(x)$).

\subsection{MMMU Negative-Control Results}\label{app:mmmu_ibp_details}

\cref{tab:mmmu_negative_control} gives the counts and scores behind \cref{subsec:ibp_mmmu}.

\begin{table}[!htbp]
    \centering
    \caption{\textbf{Pruning MMMU by $\Delta s(x)$ behaves like random removal.} LLaVA-OneVision-7B accuracy averaged over questions; gaps use unrounded scores.}\label{tab:mmmu_negative_control}
    \small
    \begin{tabular}{lcccc}
        \toprule
        Subset & $n$ & Vis.\ & Blind & Gap \\
        \midrule
        Original & 900 & 49.4 & 42.8 & 6.7 \\
        $\Delta s(x)$-ranked removal & 800 & 47.9 & 40.4 & 7.5 \\
        Random removal & 800 & -- & -- & $6.69\pm0.58$ \\
        \bottomrule
        \multicolumn{5}{l}{\scriptsize Random removal: mean $\pm$ s.d.\ over 1,000 seeds; range 4.62 to 8.75.}
    \end{tabular}
\end{table}

\subsection{VSI-Bench-Debiased}\label{app:vsi_debiasing_details}

VSI-Bench-Debiased uses hand-written per-type filters and hand-set budgets. All ten types remain, but removal rates range from 2.3\% (easy relative direction) to 85.7\% (medium relative direction; \cref{tab:vsi_debiased_statistics}). The resulting shift in question-type mix is a limitation of the pilot and motivates an automated successor.

\begin{table}[htbp]
    \centering
    \caption{\textbf{VSI-Bench-Debiased removes questions unevenly across types.}}\label{tab:vsi_debiased_statistics}
    \small
    \begin{tabular}{llrrrr}
        \toprule
        Question Type & Format & Original & Removed & Removed (\%) & Kept \\
        \midrule
        Object size & NUM & 953 & 600 & 63.0 & 353 \\
        Absolute distance & NUM & 834 & 400 & 48.0 & 434 \\
        Relative distance & MC & 710 & 400 & 56.3 & 310 \\
        Appearance order & MC & 618 & 300 & 48.5 & 318 \\
        Object counting & NUM & 565 & 314 & 55.6 & 251 \\
        Relative direction (med.) & MC & 378 & 324 & 85.7 & 54 \\
        Relative direction (hard) & MC & 373 & 257 & 68.9 & 116 \\
        Relative direction (easy) & MC & 217 & 5 & 2.3 & 212 \\
        Room size & NUM & 288 & 88 & 30.6 & 200 \\
        Route planning & MC & 194 & 80 & 41.2 & 114 \\
        \midrule
        Total & & 5,130 & 2,768 & 54.0 & 2,362 \\
        \bottomrule
    \end{tabular}
\end{table}

The pilot widens the gap more than random removal at the same rate (\cref{tab:vsi_random_control}). Across seven models (\cref{tab:vsi_transfer}), it widens most for Cambrian-S and the fine-tuned LLaVA-Video, the two with the highest original blind scores. It also widens for InternVL3-9B (11.1 to 13.2), whose language model is not from the Qwen family, so the widening is not limited to Qwen-based models. The gap barely moves for models with the lowest original blind scores.

\begin{table}[htbp]
    \centering
    \caption{\textbf{Random removal at the pilot's rate barely changes the gap.} Gap on the full set, on the debiased set, and after randomly removing 54\% of questions (mean $\pm$ s.d.\ over 10 seeds).}\label{tab:vsi_random_control}
    \small
    \begin{tabular}{lrrr}
        \toprule
        Model & Full gap & Pilot & Random \\
        \midrule
        LLaVA-Video-7B + FT & 12.4 & 16.7 & $12.4\pm0.6$ \\
        Cambrian-S & 19.2 & 26.0 & $19.4\pm0.8$ \\
        \bottomrule
    \end{tabular}
\end{table}

\begin{table}[htbp]
    \centering
    \caption{\textbf{The gap widens most for Cambrian-S and the fine-tuned LLaVA-Video-7B.} V and B are vision and blind scores (MC accuracy and NUM MRA averaged over questions; NUM answers normalized for number words and units), sorted by original blind score. Gaps use unrounded scores. Cambrian-S is a pre-release checkpoint; base LLaVA-Video-7B comes from a separate evaluation, so its scores differ slightly from \cref{tab:vsi_robust_results}.}\label{tab:vsi_transfer}
    \small
    \setlength{\tabcolsep}{3pt}
    \begin{tabular}{llrrrrrrr}
        \toprule
        & & \multicolumn{3}{c}{Original} & \multicolumn{3}{c}{Debiased} & \\
        \cmidrule(lr){3-5}\cmidrule(lr){6-8}
        Model & LLM backbone & V & B & Gap & V & B & Gap & $\Delta$ gap \\
        \midrule
        Cambrian-S & Qwen2.5 & 66.1 & 46.9 & 19.2 & 58.3 & 32.3 & 26.0 & +6.8 \\
        LLaVA-Video-7B + FT & Qwen2 & 57.1 & 44.7 & 12.4 & 48.7 & 32.0 & 16.7 & +4.3 \\
        InternVL3-9B & InternLM3 & 43.6 & 32.5 & 11.1 & 41.5 & 28.2 & 13.2 & +2.2 \\
        InternVL2.5-26B & InternLM2.5 & 45.3 & 32.5 & 12.8 & 38.4 & 25.1 & 13.3 & +0.5 \\
        InternVL2.5-8B & InternLM2.5 & 44.0 & 31.6 & 12.4 & 37.5 & 24.2 & 13.3 & +0.9 \\
        LLaVA-OneVision-7B & Qwen2 & 35.9 & 27.0 & 9.0 & 29.6 & 21.2 & 8.5 & $-0.5$ \\
        LLaVA-Video-7B & Qwen2 & 37.0 & 25.9 & 11.2 & 31.9 & 20.2 & 11.6 & +0.5 \\
        \bottomrule
    \end{tabular}
\end{table}

\subsection{Automated IBP on VSI-Bench}\label{app:auto_ibp}

Per-format IBP ranks MC and NUM questions separately because their score scales differ; the per-type variants are defined in \cref{subsec:mitigation_vsi_robust}. \Cref{tab:vsi_provenance} gives the resulting gaps.

\begin{table}[htbp]
    \centering
    \caption{\textbf{Automated per-type IBP widens the gap less than the pilot, and per-format IBP leaves it unchanged.} Gap of LLaVA-Video-7B fine-tuned on VSI-Train-10k, averaged over questions. Random removal at the same rates: 12.2 (per-format), 12.2 (per-type, uniform), 12.4 (per-type, pilot budgets).}\label{tab:vsi_provenance}
    \small
    \begin{tabular}{lrrr}
        \toprule
        Selection & Removed (\%) & Kept & Gap \\
        \midrule
        Original & 0.0 & 5,130 & 12.4 \\
        Per-format, $B{=}1000$ & 19.5 & 4,130 & 12.4 \\
        Per-type, uniform & 54.0 & 2,360 & 14.1 \\
        Per-type, pilot budgets (seeds 1, 42) & 54.0 & 2,362 & 14.5, 14.7 \\
        Pilot & 54.0 & 2,362 & 16.7 \\
        \bottomrule
    \end{tabular}
\end{table}

For per-format IBP, mean $s(x)$ falls steadily for MC and NUM as the budget grows (\cref{tab:ibp_budget_sweep}), although at $B{=}1000$ the gap does not move (\cref{tab:vsi_provenance}). We use $B{=}1000$ to balance bias reduction against coverage.

\begin{table}[htbp]
    \centering
    \caption{\textbf{Mean $s(x)$ falls steadily as the per-format IBP budget grows.} Mean TsT-RF $s(x)$ after pruning.}\label{tab:ibp_budget_sweep}
    \small
    \begin{tabular}{rrrr}
        \toprule
        & & \multicolumn{2}{c}{Mean $s(x)$} \\
        \cmidrule(lr){3-4}
        $B$ & Removed (\%) & MC & NUM \\
        \midrule
        200 & 3.9 & 0.337 & 0.421 \\
        500 & 9.7 & 0.308 & 0.385 \\
        1,000 & 19.5 & 0.270 & 0.329 \\
        1,500 & 29.2 & 0.246 & 0.270 \\
        2,000 & 39.0 & 0.219 & 0.210 \\
        2,500 & 48.7 & 0.200 & 0.138 \\
        \bottomrule
    \end{tabular}
\end{table}

\subsection{Diagnostic Dependence}\label{app:diag_dependence}

TsT-LLM and TsT-RF remove largely different questions (\cref{tab:cross_diagnostician}): their removed sets have Jaccard 0.060 (68 shared of 1,132 in the union). Part of the difference may come from their ranking scores, $\Delta s(x)$ for TsT-LLM and $s(x)$ for TsT-RF.

\begin{table}[htbp]
    \centering
    \caption{\textbf{TsT-LLM and TsT-RF remove most question types at very different rates.} Removal rates (\%) by question type. Budgets are 200 for TsT-LLM and 1,000 for TsT-RF. At a matched budget of 200, Jaccard is 0.050.}\label{tab:cross_diagnostician}
    \small
    \begin{tabular}{lrr}
        \toprule
        Question type & TsT-LLM & TsT-RF \\
        \midrule
        Appearance order & 15.5 & 40.1 \\
        Object counting & 15.9 & 15.2 \\
        Object size & 0.4 & 31.2 \\
        Absolute distance & 0.2 & 12.2 \\
        \bottomrule
    \end{tabular}
\end{table}

Replacing Qwen2-7B with LLaMA-3-8B in TsT-LLM (same folds and seed, native chat templates, no room-size questions) gives $\Delta s(x)$ rankings with Spearman 0.58 and 59.7\% top-1,000 overlap. Within TsT-RF, forests of 200 to 2,000 trees and depth 10, 20, or unlimited give pairwise Spearman at least 0.990 and top-1,000 Jaccard at least 0.873. Rankings thus change substantially across base models but little across forest settings; there is no single correct pruned set.

\FloatBarrier
\section{TsT-RF Interpretability Analysis}\label{app:rf_interpretability}

We illustrate TsT-RF on VSI-Bench's object-size questions (\texttt{object\_size\_estimation}). Feature importances explain TsT-RF; they are not a faithful explanation of how an MLLM answers.

\paragraph{Setup.}
The question template is ``What is the length of the longest dimension (length, width, or height) of the \{object\}, measured in centimeters?''
Each question gets six features: the object category and five statistics of that category computed on the training folds only (\texttt{obj\_count}, \texttt{obj\_freq\_score}, \texttt{obj\_val\_log\_mean}, \texttt{obj\_val\_log\_std}, \texttt{obj\_val\_log\_ratio}).

\paragraph{Results.}
With 5-fold cross-validation, TsT-RF reaches 61.4\% MRA, well above VSI-Bench's 29.9\% frequency baseline for this task~\citep{yang2024think}.
Gini importance concentrates on one feature, \texttt{obj\_val\_log\_mean} (the category's mean log size), at 0.968; the other five together contribute less than 0.04. The forest thus mostly predicts each category's typical size.

\paragraph{Root Cause.}
Categories with very low size variation, such as dishwashers, beds, and washers, can be answered from category statistics alone; high-variation categories such as ceiling lights and radiators are much harder to predict this way (\cref{tab:high_sx_size_examples}).

\begin{table}[!htbp]
    \centering
    \caption{\textbf{Categories with near-constant sizes are easy to exploit.} Size statistics for selected categories. CV is the coefficient of variation (standard deviation over mean).}
    \label{tab:high_sx_size_examples}
    \begin{tabular}{lccc}
        \toprule
        Object & Mean (cm) & Std Dev (cm) & CV \\
        \midrule
        \multicolumn{4}{l}{\textit{Low-variance objects (easily exploitable)}} \\
        \midrule
        Dishwasher & 90.4 & 3.4 & 0.037 \\
        Bed & 216.1 & 17.2 & 0.080 \\
        Washer & 87.1 & 5.0 & 0.058 \\
        Kettle & 23.8 & 1.5 & 0.062 \\
        Mouse & 11.6 & 1.1 & 0.091 \\
        \midrule
        \multicolumn{4}{l}{\textit{High-variance objects (harder to exploit)}} \\
        \midrule
        Ceiling Light & 71.8 & 55.8 & 0.778 \\
        Radiator & 146.3 & 110.4 & 0.755 \\
        \bottomrule
    \end{tabular}
\end{table}

\paragraph{Implication for Benchmark Design.}
To close this shortcut, designers can remove questions about low-variance categories or sample instances so that sizes vary within each category.

\FloatBarrier

\end{document}